\documentclass{article}

\PassOptionsToPackage{numbers, compress}{natbib}

\usepackage{graphicx}
\usepackage{amsmath}
\usepackage{subcaption}
\usepackage[preprint]{neurips_2024}

\usepackage[utf8]{inputenc} 
\usepackage[T1]{fontenc}    
\usepackage{hyperref}       
\usepackage{url}            
\usepackage{booktabs}       
\usepackage{amsfonts}       
\usepackage{nicefrac}       
\usepackage{microtype}      
\usepackage{xcolor}         
\usepackage{multirow}

\usepackage[capitalize,noabbrev]{cleveref}

\newcommand{\icotsi}{{ICoT-SI}}
\newcommand{\icotkd}{{ICoT-KD}}

\title{From Explicit CoT to Implicit CoT: \\Learning to Internalize CoT Step by Step}

\author{%
Yuntian Deng$^{1,2}$ \quad  Yejin Choi$^{1,3}$ \quad Stuart Shieber$^4$\\
$^{1}$Allen Institute for Artificial Intelligence \quad $^{2}$University of Waterloo \\ $^{3}$University of Washington \quad
$^{4}$Harvard University\\
\texttt{\{yuntiand, yejinc\}@allenai.org}, \texttt{shieber@seas.harvard.edu} }

\begin{document}

\maketitle

\begin{abstract}
When leveraging language models for reasoning tasks, generating explicit chain-of-thought (CoT) steps often proves essential for achieving high accuracy in final outputs. In this paper, we investigate if models can be taught to internalize these CoT steps. To this end, we propose a simple yet effective method for internalizing CoT steps: starting with a model trained for explicit CoT reasoning, we gradually remove the intermediate steps and finetune the model. This process allows the model to internalize the intermediate reasoning steps, thus simplifying the reasoning process while maintaining high performance. Our approach enables a GPT-2 Small model to solve 9-by-9 multiplication with up to 99\% accuracy, whereas standard training cannot solve beyond 4-by-4 multiplication. Furthermore, our method proves effective on larger language models, such as Mistral 7B, achieving over 50\% accuracy on GSM8K without producing any intermediate steps.
\end{abstract}

\section{Introduction}
A prevalent approach to improving the performance of language models (LMs) to perform complex reasoning tasks is chain-of-thought (CoT) reasoning, in which the LM generates explicit intermediate reasoning steps before arriving at a final answer~\citep{nye2021work,wei2022chain}. This method allows models to break down complex problems into simpler, manageable parts, thereby improving the accuracy of their final predictions. However, this explicit reasoning process can be computationally expensive, especially when the reasoning chain is long~\citep{deng2023implicit}. Additionally, using explicit intermediate steps might not align with the intrinsic computational strengths of LMs~\citep{lehnert2024a}: for instance, multi-digit multiplication is very easy for calculators but remains challenging for GPT-4~\citep{yang2023gpt}.

In this work, we examine the possibility of internalizing the reasoning process in the model's hidden states. We propose an approach, Stepwise Internalization, which begins with a model trained for explicit CoT reasoning. We then gradually remove the intermediate steps and finetune the model, forcing it to internalize the reasoning process. Once all intermediate steps are internalized, we achieve a model capable of full implicit CoT reasoning. Moreover, even in cases where the model does not have the capacity for full implicit CoT reasoning, this method still allows for shortening the reasoning chain while maintaining accuracy.

Our approach is an alternative to the approach proposed by \citet{deng2023implicit}, which shares the goal of implicitly reasoning using the hidden states of transformers instead of relying on explicit CoT tokens. To teach the model to use hidden states for reasoning, that method employs a teacher model that performs explicit CoT reasoning, and then distills the teacher's hidden states into the student model's hidden states. In comparison, our approach is much simpler yet more effective. 

Our approach demonstrates significant improvements over standard training methods. For instance, a GPT-2 Small model trained with Stepwise Internalization on multiplication can solve even 9-by-9 multiplication problems nearly perfectly, while standard training without CoT struggles even with 4-by-4 multiplication. Furthermore, our method scales effectively to larger models, such as the Mistral 7B model~\citep{jiang2023mistral}, achieving over 50\% accuracy on the GSM8K dataset of grade-school math word problems~\citep{cobbe2021training}, without producing any explicit intermediate steps, outperforming the much larger GPT-4 model without chain-of-thought reasoning, which only scores 44\% when prompted to directly generate the answer.

It is important to note that our empirical evaluation focuses on specific reasoning tasks like multi-digit multiplication and grade-school math problems. While our results show the potential of Stepwise Internalization in these contexts, and the simplicity of the method makes it applicable to chain-of-thought approaches in a wide range of tasks, further research is needed to explore its efficacy across a broader range of tasks and more diverse CoT traces. Due to limitations in available computational resources, experiments on other tasks are out of scope for this work. This paper aims to lay the groundwork for this new approach and highlight its promise, while acknowledging that its full generalization is still under investigation.

The contributions of our work are as follows: First, we introduce Stepwise Internalization, a simple method for implicit CoT reasoning. Second, we demonstrate the effectiveness of internalizing intermediate hidden states via Stepwise Internalization. Third, we provide empirical results showing the superior performance of models trained with Stepwise Internalization on different reasoning tasks and model scales. Our code, data, and pretrained models are available at \url{https://github.com/da03/Internalize_CoT_Step_by_Step}.

\section{Background: Implicit Chain-of-Thought Reasoning}
Implicit chain-of-thought reasoning (implicit CoT, or ICoT) is a concept introduced by \citet{deng2023implicit}, where during generation, the language model does not produce explicit intermediate reasoning steps in words. It is distinct from not using chain-of-thought reasoning (No CoT), in that explicit reasoning steps are allowed during training, enabling the ICoT model to learn the underlying reasoning approach from the supervision provided on the reasoning process. The key insight of \citet{deng2023implicit} is that intermediate reasoning steps serve two purposes in explicit CoT: they provide supervision during training to facilitate learning the task~\citep{nye2021work}, and they act as a scratchpad during inference to assist in solving the task~\citep{wei2022chain}. However, the latter purpose can be fulfilled by utilizing the internal states of the model instead of explicit tokens.

As an illustrative example, consider using a language model to solve a multi-digit multiplication problem, such as $12 \times 34$. (The actual input reverses the digit order as \verb|2 1 * 4 3| for consistency with \citet{deng2023implicit}.) In the long multiplication algorithm, $12 \times 34$ is broken into:
\begin{equation*}
12 \times 4 + 12 \times 30 = \underbrace{48}_{\text{reversed: }84} + \underbrace{360}_{\text{reversed: }063}.
\end{equation*}
In explicit CoT, the model is trained to predict these intermediate steps \verb|8 4 + 0 6 3| before predicting the final answer \verb|8 0 4| ($408$ reversed). Predicting these intermediate steps facilitates the model's ability to solve the task. (The intermediate steps are also reversed to make it easier for the model to predict~\citep{shen2023positional}.) 

In both No CoT and implicit CoT settings, the model needs to directly predict the answer $408$ from the input, bypassing the intermediate steps. This approach can make inference much faster for long reasoning chains, albeit at the cost of accuracy.

The primary difference between implicit CoT and No CoT lies in the use of intermediate reasoning steps as supervision during training. In the work of \citet{deng2023implicit}, a knowledge distillation approach was employed to distill explicit reasoning into implicit reasoning within the hidden states. This method involves training a teacher model to perform explicit CoT reasoning and then transferring this knowledge to a student model, which internalizes the reasoning process within its hidden states.

In the present work, we propose a far simpler yet more effective approach based on a kind of curriculum learning that we call Stepwise Internalization, which we detail in the next section.

\begin{figure}[!t]
  \centering
  \includegraphics[width=0.95\textwidth]{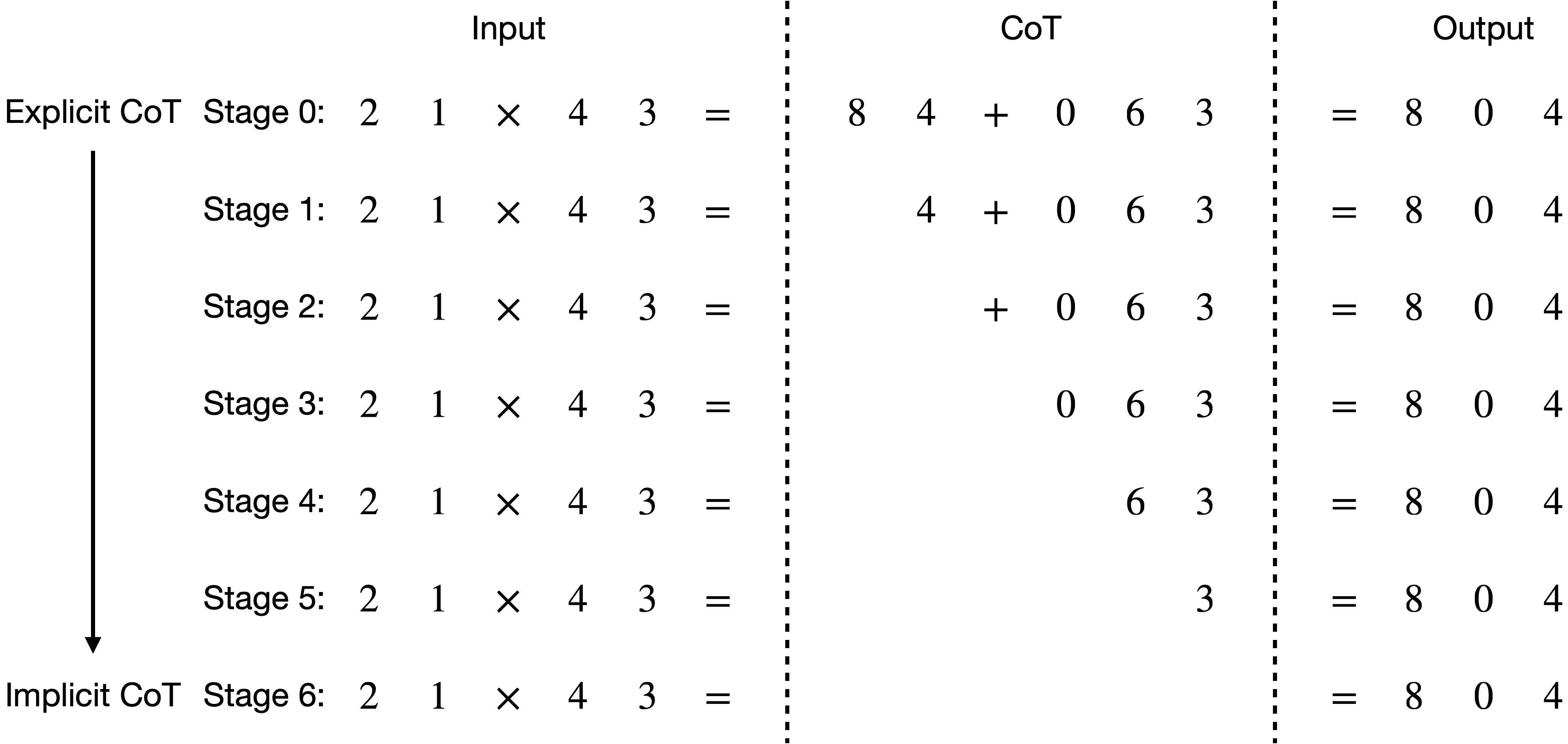}
  \caption{\label{fig:main}Stepwise Internalization for Implicit Chain-of-Thought Reasoning. This figure illustrates the Stepwise Internalization method using the example of solving $12 \times 34$. The training process consists of multiple stages. At Stage 0, the model is trained to predict both the full chain-of-thought (CoT) and the final output, which is the same as explicit CoT training. At Stage 1, the first CoT token is removed, and the model is finetuned to predict the remaining CoT tokens and the output. This process continues with each subsequent stage removing an additional CoT token. By Stage 6, all CoT tokens have been removed, and the model is trained to directly predict the output from the input, achieving implicit CoT reasoning. This gradual removal and finetuning process allows the model to gradually internalize the reasoning steps.}
\end{figure}

\section{\label{sec:method}Stepwise Internalization}
Stepwise Internalization is a method designed to achieve implicit chain-of-thought reasoning by gradually removing intermediate reasoning steps during training. We define the input as $x$, the intermediate steps as $z=z_1,z_2,\cdots,z_m$, and the final output as $y$. A language model with parameters $\theta$ is first trained using the following loss function:
\begin{equation*}
\min_\theta -\log P_\theta (y, z_{1:m} \mid x),
\end{equation*}
where $z_{1:m}$ denotes the sequence of intermediate steps $z_1,z_2,\cdots,z_m$.

At each step $t$ of the training process, we remove (up to) $s(t)$ tokens from the intermediate steps $z$. The updated loss function then becomes:
\begin{equation*}
\min_\theta -\log P_\theta (y, z_{1+\min(s(t), m):m} \mid x).
\end{equation*}

There are multiple ways to parameterize $s(t)$. For instance, it might be based on a threshold of the loss value or a predefined schedule similar to learning rate schedulers used in optimizers. In this work, for simplicity, we use a linear schedule for removing tokens:
\begin{equation*}
s(t) = \left\lfloor \Delta \frac{t}{T} \right\rfloor,
\end{equation*}
where $T$ is the total number of steps per epoch, and $\Delta$ is a hyperparameter controlling how many CoT tokens are removed per epoch. (Once $s(t)$ exceeds the number of actual chain-of-thought tokens, all tokens are removed.)

During initial experiments, we observed instability in the training process due to changes in the loss function over time. This instability arises for two primary reasons:

First, the optimizer commonly used in training language models, such as AdamW~\citep{kingma2017adam,loshchilov2018decoupled}, maintains estimates of second-order gradients. A sudden change in the loss function, caused by the removal of one more CoT token, results in abrupt changes in the second-order gradients. To address this issue, we reset the optimizer's state whenever an additional CoT token is removed.

Second, even if a model fits perfectly to the current loss when $s$ tokens are removed, transitioning to the next stage, where $s+1$ tokens are removed, leads to a significant increase in the loss, as the model is not yet trained for this new setting. To mitigate this issue, we introduce a technique which we term ``Removal Smoothing'', where we add a small random offset to the original number of tokens to remove $s(t)$, such that:
\begin{equation*}
s(t)^* = s(t) + o,
\end{equation*}
where $o$ is a random variable with support of non-negative integers $\mathbb{Z}_{\ge0}$, and its distribution is parameterized by another hyperparameter $\lambda$:
\begin{equation*}
P(o) \propto \exp(-\lambda o).
\end{equation*}
When $\lambda=\infty$, $o=0$ and we recover the version without Removal Smoothing. However, when $\lambda<\infty$, the model is trained to remove more than $s(t)$ tokens at step $t$ with a small probability, which helps smooth the transition into the next stage of removing $s(t)+1$ tokens, reducing the abrupt jumps in the loss function.

\Cref{fig:main} illustrates the high-level idea of the Stepwise Internalization approach. The training process consists of multiple stages, where the model progressively learns to internalize reasoning steps by removing tokens from the CoT at each stage, eventually achieving implicit CoT reasoning.

\begin{table*}[!t]
    \centering
    \caption{\label{tab:dataset}Dataset statistics. The number of tokens is the median based on the GPT-2 tokenizer.}
    \begin{tabular}{lc@{ }c@{ }ccc@{ }c@{ }ccc@{ }c@{ }ccc@{ }c@{ }c}
    \toprule
    \multirow{2}{*}{Dataset} & \multicolumn{3}{c}{Size} & &\multicolumn{3}{c}{\# Input Tokens}               & &\multicolumn{3}{c}{\# CoT Tokens} & &\multicolumn{3}{c}{\# Output tokens}  \\    
    \cmidrule{2-4}\cmidrule{6-8}\cmidrule{10-12}
   \cmidrule{14-16}
   &  Train  & Dev & Test  &  &  Train & Dev & Test   && Train & Dev & Test && Train & Dev & Test   \\
    \midrule
        $4\times4$ Mult & 808k & 1k & 1k && 9 & 9 & 9 && 46 & 46 & 46 && 9 & 9 & 9 \\
        $5\times5$ Mult & 808k & 1k & 1k && 11 & 11 & 11 && 74 & 74 & 74 && 11 & 11 & 11 \\
        $7\times7$ Mult & 808k & 1k & 1k && 15 & 15 & 15 && 148 & 148 & 148 && 15 & 15 & 15 \\
        $9\times9$ Mult & 808k & 1k & 1k && 19 & 19 & 19 && 246 & 246 & 246 && 19 & 19 & 19 \\
        GSM8K & 378k & 0.5k & 1.3k && 40 & 51 & 53 && 19 & 21 & 24 && 2 & 2 & 2\\
        \bottomrule
    \end{tabular}
\end{table*}

\section{\label{sec:experiments}Experimental Setup}
\subsection{Data}
We evaluate our proposed Stepwise Internalization method on two reasoning tasks following \citet{deng2023implicit}: multi-digit multiplication and grade-school math reasoning.

\paragraph{Multi-digit multiplication.} We use two of the most challenging arithmetic tasks from BIG-bench \citep{srivastava2023beyond}: 4-by-4 multiplication and 5-by-5 multiplication, as described by \citet{deng2023implicit}. Given the effectiveness of Stepwise Internalization on these tasks, we extend our evaluation to 7-by-7 and 9-by-9 multiplication. The complexity of multiplication tasks grows significantly with the number of digits, as the program length grows quadratically with the number of digits~\citep{dziri2024faith}. We use the scripts and setup from \citet{deng2023implicit} to generate synthetic training data for our main experiments\footnote{Following \citet{deng2023implicit}, $K$-by$K$ multiplication only considers $K$-digit numbers but not lower digits.}.

\paragraph{Grade school math.} We use the GSM8K dataset~\citep{cobbe2021training}, with the augmented training data provided by \citet{deng2023implicit}. Detailed dataset statistics are provided in \Cref{tab:dataset}.

\subsection{Baselines and Models}
We compare our method to the following baselines:
\begin{itemize}
    \item \textbf{No CoT:} Models directly trained without chain-of-thought supervision.
    \item \textbf{Explicit CoT:} Models finetuned or prompted with explicit chain-of-thought reasoning~\citep{nye2021work}. We use 5-shot prompting for GPT 3.5 and GPT-4 but full finetuning for other models.
    \item \textbf{\icotkd:} The implicit chain-of-thought via knowledge distillation method proposed by \citet{deng2023implicit}.
\end{itemize}

Our proposed method, implicit chain-of-thought via Stepwise Internalization, is termed \icotsi. To verify the effectiveness of our approach across different model scales, we use pretrained models GPT-2~\citep{radford2019language}, Phi-3 3.8B~\citep{abdin2024phi3}, and Mistral-7B~\citep{jiang2023mistral}.

\subsection{Evaluation}
Because the premise for implicit chain-of-thought methods is to approach the speed of no chain-of-thought and the accuracy of explicit chain-of-thought, we use two main evaluation metrics: First, we evaluate the accuracy of each method on the respective tasks of generating the final output. Second, we compare the inference speed of each method to the No CoT baseline. We measure speed, in examples per second, on an Nvidia H100 GPU with a batch size of 1. For \icotkd, we directly take numbers from \citet{deng2023implicit}. However, due to hardware differences, we recompute speed relative to No CoT when speed numbers from \icotkd{} are not available.


\begin{table*}[!t]
    \centering
    \caption{\label{tab:main}Results on multiplication tasks. \icotkd: Implicit CoT via knowledge distillation, with numbers taken from \citet{deng2023implicit}. \icotsi: Implicit CoT via Stepwise Internalization (this work). Accuracy (Acc in the table) measures the exact match accuracy of producing the final answer. Speed measures the number of examples per second during inference using a batch size of 1, normalized by the speed of the corresponding No CoT model. In addition to the GPT2-Small model on which our \icotsi{} experiments were based, for comparison purposes we provide a set of No CoT and explicit CoT baselines for a set of models of wide-ranging sizes. $^\dagger$: 5-shot prompted instead of finetuned.} 
    \begin{tabular}{@{}lccccccccccc@{}}
    \toprule
    Model & \multicolumn{2}{c}{$4\times4$} & &\multicolumn{2}{c}{$5\times5$}               & &\multicolumn{2}{c}{$7\times7$} & &\multicolumn{2}{c}{$9\times9$}\\
    \cmidrule{2-3}\cmidrule{5-6}\cmidrule{8-9}
   \cmidrule{11-12} 
   &  Acc  & Speed &     &  Acc & Speed &    & Acc & Speed & & Acc & Speed\\
    \midrule
    \textbf{GPT-2 Small (117M)}\\
     \ \ Explicit CoT & 1.00 & 0.17 & &1.00 & 0.14 & & 1.00 &0.12&& 1.00& 0.09 \\
    \ \ No CoT & 0.29 & 1.00 & & 0.01 & 1.00 & & 0.00 &1.00&& 0.00&1.00\\
    \ \ \icotkd & 0.97& 0.67 &  & 0.10 & 0.71 & & - & - &&-&-\\
    \ \ \icotsi  & 1.00 & 1.02 & & 0.95 & 1.00 & & 0.95 & 1.00 & & 0.99 & 1.00 \\
    \midrule
    \textbf{MathGLM-100M}\\
    \ \ No CoT & 0.80 & 1.00 & & 0.56 & 1.00 & &-&-&&-&-\\
    \midrule
    \textbf{MathGLM-500M}\\
    \ \ No CoT & 0.90 & 1.00 & & 0.60 & 1.00 & &-&-&&-&-\\
    \midrule
    \textbf{MathGLM-2B}\\
    \ \ No CoT & 0.95 & 1.00 & & 0.90 & 1.00 & &-&-&&-&-\\
    \midrule
    \textbf{GPT-3.5}$^\dagger$\\
    \ \ Explicit CoT & 0.43 & 0.10 & & 0.05 & 0.07 & &0.00&0.15&&0.00&0.11\\
    \ \ No CoT & 0.02 & 1.00 & & 0.00 & 1.00 & & 0.00& 1.00&& 0.00 & 1.00\\
    \midrule
    \textbf{GPT-4}$^\dagger$\\
    \ \ Explicit CoT  & 0.77 & 0.14 & & 0.44 & 0.14 & &0.03&0.09&&0.00&0.07 \\
    \ \ No CoT & 0.04  & 1.00 & & 0.00 & 1.00 & &0.00 & 1.00&&0.00&1.00 \\
    \bottomrule
    \end{tabular}
\end{table*}
\begin{table*}[!t]
    \centering
    \caption{\label{tab:main2}Accuracy of various approaches on GSM8K. $^\dagger$: 5-shot prompted instead of finetuned.} 
    \begin{tabular}{@{}lcccccc@{}}
    \toprule
    Model & GPT-2 Small & GPT-2 Medium & Phi-3 3.8B & Mistral 7B & GPT-3.5$^\dagger$ & GPT-4$^\dagger$ \\    
    \midrule
    Explicit CoT & 0.41 & 0.44 & 0.74 & 0.68 & 0.62 & 0.91\\
    \midrule
    No CoT & 0.13 & 0.17 & 0.28 & 0.38 & 0.03 & 0.44 \\
    \icotkd & 0.20 & 0.22 & - & - & - & -\\
   \icotsi & 0.30 & 0.35 & 0.31 & 0.51 & - & - \\
    \bottomrule
    \end{tabular}
\end{table*}
\section{Results}
\Cref{tab:main} presents the main results, where we compare Stepwise Internalization to various baselines.

\begin{figure}[!t]
  \centering
  \begin{subfigure}[b]{0.73\textwidth}
  \centering
  \includegraphics[height=7.8cm]{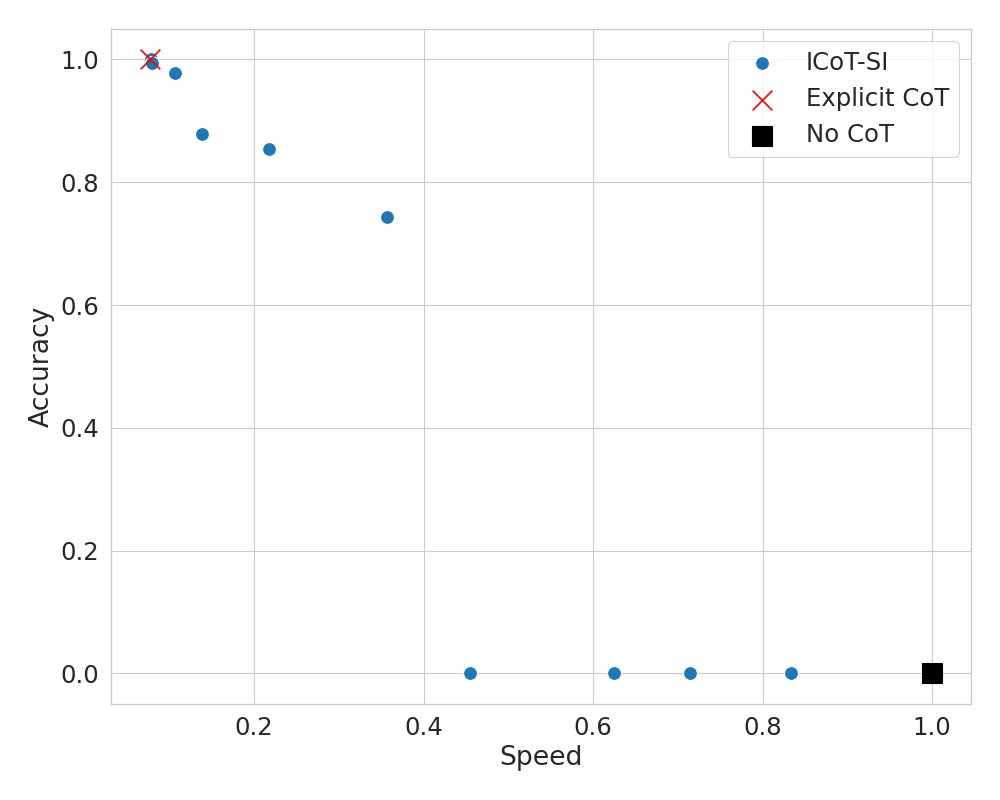}
  \caption{\label{fig:tradeoff}}
  \end{subfigure}
  \begin{subfigure}[b]{0.25\textwidth}
  \centering
  \includegraphics[height=7.8cm]{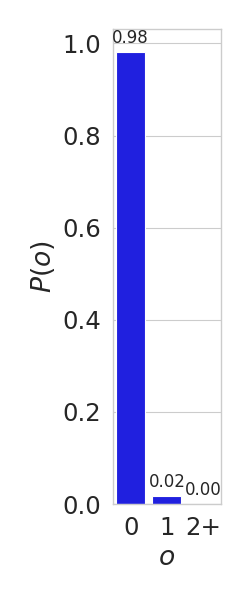}
  \caption{\label{fig:lamb}}
  \end{subfigure}
  \caption{(a) Trade-off of Speed versus Accuracy for \icotsi. This figure illustrates the trade-off between speed and accuracy enabled by the Stepwise Internalization approach (\icotsi) for the $11\times11$ multiplication task using GPT-2 Small. As more CoT tokens are removed and internalized, accuracy decreases while speed increases. At the two extremes, our approach can recover both Explicit CoT (high accuracy but slow) and No CoT (very fast but with an accuracy of 0). Note that we removed points that are ``dominated'' by other points (i.e., there exists another point with both higher speed and higher accuracy) in this figure.
 (b) Distribution over Random Removal Offset $o$ in Removal Smoothing with $\lambda=4$. The distribution is mostly concentrated at $o=0$ with a probability of 0.98, and $o\ge1$ has a probability of only 0.02. Despite this, the removal smoothing proves to be effective, as demonstrated in the ablation studies. }
\end{figure}
\paragraph{Stepwise Internalization is effective.}
Compared to other methods that do not output intermediate steps, Stepwise Internalization (\icotsi) proves to be highly effective. For example, \icotsi{} enables a GPT-2 Small model to solve the $9\times9$ multiplication problem with an accuracy of 0.99, whereas the No CoT method fails on even $4\times4$ multiplication. Additionally, \icotsi{} outperforms Implicit CoT via Knowledge Distillation (\icotkd); while \icotkd{} fails to solve $5\times5$ multiplication using a GPT-2 Small model, \icotsi{} can solve up to $9\times9$ multiplication. Also, while \icotkd{} is slightly slower than No CoT due to the additional emulator model, \icotsi{} has the same speed as No CoT\footnote{The speed of \icotsi{} in \Cref{tab:main} is not always 1.00 due to randomness in hardware speed.}.

When compared to existing literature, \icotsi{} is also competitive. For instance, at a similar model size, MathGLM-100M~\citep{yang2023gpt} can only solve $5\times5$ multiplication with an accuracy of 0.56. Even with 2 billion parameters, MathGLM-2B can solve $5\times5$ multiplication with an accuracy of 0.90. Although another related work \citep{shen2023positional} is able to train a GPT-2 Small model to solve up to $14\times14$ multiplication, the method proposed in that work is specific to arithmetic tasks, whereas \icotsi{} is more general.

\icotsi{} enables the internalization of CoT reasoning in a general way, making it applicable to tasks beyond arithmetic, such as grade-school math problems. For example, on the GSM8K dataset, \icotsi{} achieves a new state-of-the-art accuracy for models not using any intermediate steps. It finetunes the Mistral-7B model to achieve over 0.50 accuracy, whereas even GPT-4 can only achieve 0.44 without using intermediate steps.

\paragraph{Stepwise Internalization lags behind explicit CoT in accuracy but is faster.}
In terms of accuracy, implicit CoT methods still lag behind explicit CoT. For instance, a finetuned Mistral-7B model can achieve an accuracy of 0.68 on GSM8K with explicit CoT but \icotsi{} only got 0.51. However, implicit CoT methods offer significant speed advantages. For example, on the $9\times9$ multiplication task, \icotsi{} is comparable in accuracy to Explicit CoT but is 11 times faster during inference.

Overall, our results demonstrate that Stepwise Internalization is an effective method for enabling implicit CoT reasoning, offering a compelling trade-off between accuracy and speed. This makes it a valuable approach for tasks requiring both high performance and low latency.


\begin{figure}[!t]
  \centering
  \includegraphics[width=0.99\textwidth]{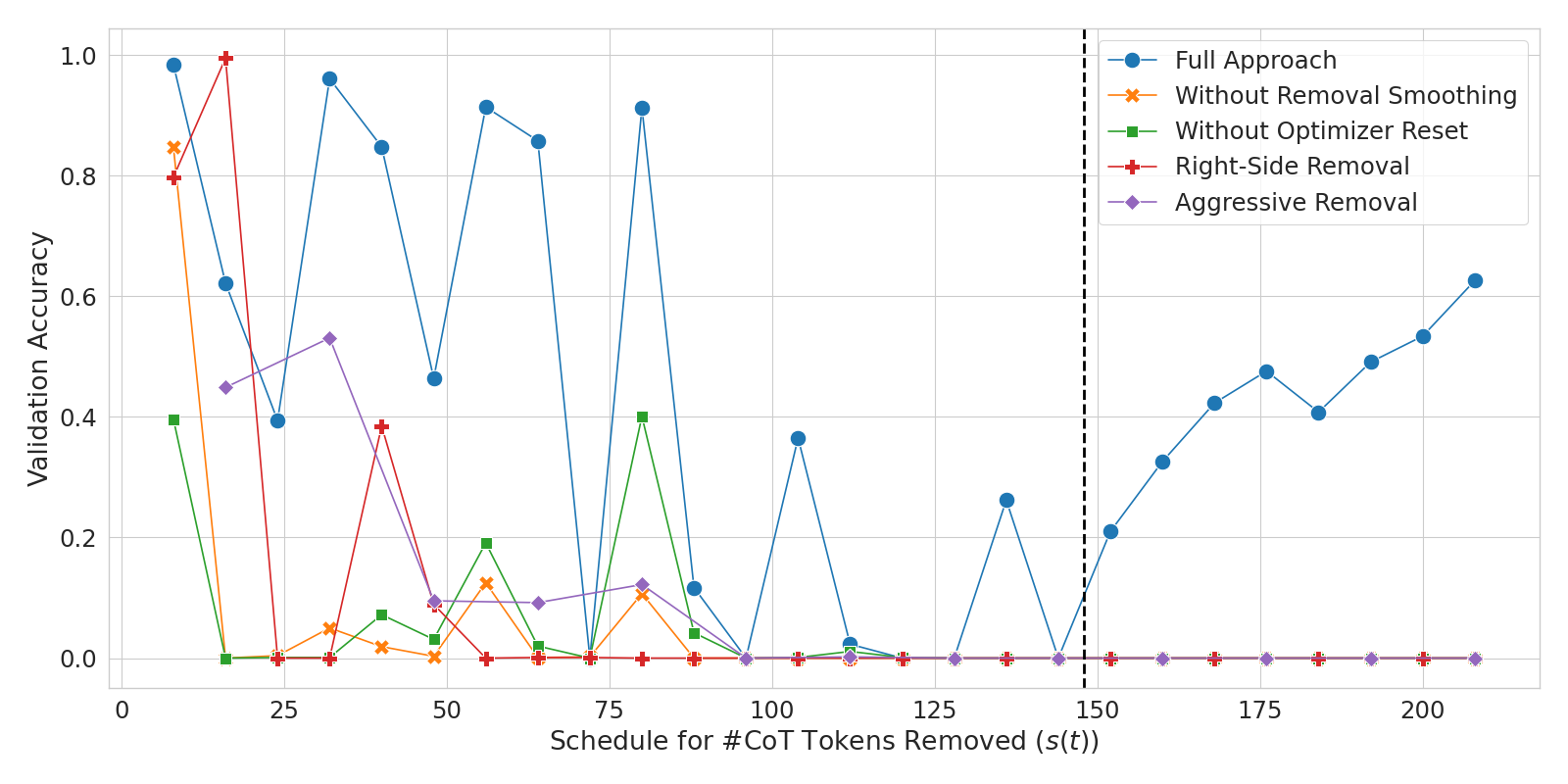}
  \caption{\label{fig:acc_removed}Accuracy during training for various ablations. This figure plots the validation accuracy as a function of the potential number of removed CoT tokens during training for the $7\times7$ multiplication task using GPT-2 Small. The black dashed vertical line indicates the point at which the schedule has removed all CoT tokens. The curves compare the following variants: ``Full Approach''; ``Without Removal Smoothing'' (removal smoothing with $\lambda=4$ is not used, i.e., $\lambda=\infty$); ``Without Optimizer Reset'' (optimizer states are not reset after removing a token); ``Right-Side Removal'' (CoT tokens are removed from the end instead of the beginning); and ``Aggressive Removal'' (16 instead of 8 CoT tokens are removed per epoch). All these variants underperform the full approach. For more details, see \Cref{sec:ablation}.}
\end{figure}

\section{Analysis}
\subsection{Accuracy-Speed Trade-off}
One notable advantage of \icotsi{} is that it allows trading off accuracy with speed by internalizing different amounts of CoT tokens. At one extreme, when no tokens are internalized, \icotsi{} can recover explicit CoT performance. At the other extreme, when all tokens are internalized, we achieve implicit CoT, typically with much better performance than directly training a No CoT model.

Even when \icotsi{} is not completely successful due to model capacity limitations, such as on more challenging tasks where it cannot internalize all CoT steps, we can still leverage intermediate checkpoints to achieve a trade-off between accuracy and speed. For example, as shown in \Cref{fig:tradeoff}, on the $11\times11$ multiplication task with GPT-2 Small, even though the model cannot internalize all CoT steps, \icotsi{} is still able to achieve an accuracy of over 0.7 at a speed four times that of explicit CoT when part of the CoT tokens are internalized.

This trade-off curve illustrates the flexibility of \icotsi{} in balancing computational efficiency and model performance. By adjusting the number of internalized CoT tokens, users can optimize for either higher accuracy or faster inference depending on the requirements of their specific application.

\subsection{\label{sec:ablation}Ablation Studies}
\Cref{fig:acc_removed} plots the validation accuracy versus the schedule for the number of CoT tokens removed during training for the $7\times7$ multiplication task. This figure compares the full approach to several ablated variants. Even for the full approach, there are fluctuations in the curve, and the validation accuracy briefly drops to zero at one point during training but eventually recovers. However, the ablated variants do not fully recover when accuracy drops.

\paragraph{Removal smoothing.}
As mentioned in \Cref{sec:method}, adding a small random offset $o$ to the number of removed tokens is crucial when the loss function changes due to the removal of more CoT tokens. The distribution of $o$ is parameterized by a hyperparameter $\lambda$, as introduced in \Cref{sec:method}. We use $\lambda=4$ throughout this work, resulting in the distribution shown in \Cref{fig:lamb}. In this distribution, 98\% of the time, $o=0$, but about 2\% of the time, one or more additional tokens are removed. As shown in \Cref{fig:acc_removed}, the ``Without Removal Smoothing'' curve fails to recover after the accuracy drops to zero at around $s(t)=50$, whereas the full approach does much better.

\paragraph{Resetting the optimizer.}
Another important technique for stabilizing training is resetting the optimizer when more tokens are removed. This avoids large estimates of second-order derivatives and stabilizes training. In \Cref{fig:acc_removed}, the ``Without Optimizer Reset'' curve drops to zero around 100 steps and does not recover, showing the importance of resetting the optimizer during training.

\paragraph{Removal side.}
In our main experiments, CoT tokens are removed from the beginning (left side). Removing CoT tokens from the right side performs significantly worse, as shown by the ``Right-Side Removal'' curve in \Cref{fig:acc_removed}. We suspect this is because internalizing tokens at the beginning is easier than internalizing tokens at the end. CoT tokens at the end depend on the earlier tokens, so internalizing them between the end of CoT and the beginning of the final answer, which only has a few positions, is more challenging. In contrast, internalizing tokens at the beginning allows distributing them across the entire input.

\paragraph{Number of tokens removed per epoch.}
The number of tokens removed per epoch ($\Delta$) significantly affects the training stability and speed. In the main experiments, we used $\Delta=8$, which removes 8 tokens per epoch. A higher $\Delta$ value leads to faster training but risks not converging, as the model may not be able to keep up with the rapid changes in the loss function. For instance, when using $\Delta=16$, the training fails to converge, as shown by the ``Aggressive Removal'' curve in \Cref{fig:acc_removed}. Conversely, a lower $\Delta$ value is more likely to result in successful training but at a slower pace. Future work could explore adaptive $\Delta$ schedules based on loss values to balance speed and stability more effectively.

\section{Related Work}
\paragraph{No CoT approaches.}
Several works in the literature focus on training language models to solve arithmetic tasks without outputting intermediate steps. MathGLM~\citep{yang2023gpt} demonstrated that with sufficient training data, including both lower-digit and higher-digit arithmetic task demonstrations, a 2 billion parameter LM can solve multi-digit arithmetic tasks without any intermediate steps. Compared to this work, Stepwise Internalization achieves higher accuracy in solving multi-digit multiplication with much smaller models, likely due to leveraging chain-of-thought supervision during training. Another notable work by \citet{shen2023positional} showed that by mixing lower-digit and higher-digit multiplication demonstrations, even a GPT-2 Small can learn up to 14-digit multiplication. However, Stepwise Internalization does not require specially prepared training data with mixed task difficulties. Additionally, Stepwise Internalization is theoretically applicable to any reasoning task with CoT reasoning steps, as demonstrated by its effectiveness on grade-school math problems.

Also relevant is the work of \citet{pfau2024lets}, which shows that transformer language models can reason using filler tokens as an alternative to CoT tokens. They showed reasoning using these filler tokens improves a language model's expressivity. Our approach has the potential to be combined with their approach to solve even more challenging tasks. 

\paragraph{Internalizing CoT.}
Our work is closely related to that of \citet{deng2023implicit} (\icotkd), which introduced the task of implicit CoT reasoning. \icotkd{} allows using CoT during training but not during generation, and it implements this via knowledge distillation to internalize the reasoning steps within hidden states. Compared to \icotkd, Stepwise Internalization has three advantages: First, it is simpler to implement as it does not require a teacher model. Second, while \icotkd{} internalizes reasoning into a single ``column'' of states (corresponding to the final input position), Stepwise Internalization allows the model to internalize reasoning across all input positions. Lastly, Stepwise Internalization achieves better accuracy compared to \icotkd.

Our work is also related to Context Distillation~\citep{snell2022learning}, which trains a model to produce the same output when conditioned on a scratchpad versus without it. Each stage of Stepwise Internalization can be viewed as a form of context distillation, where one CoT token is distilled into the model's internal states. Stepwise Internalization extends Context Distillation into a curriculum learning setting. 

Another relevant work is Searchformer~\citep{lehnert2024a}, which first trains a transformer to imitate A* search and then finetunes it on sampled shorter search traces. This allows the model to perform searches using fewer steps than those provided during training. While Searchformer relies on sampling to find shorter traces, Stepwise Internalization forces the model to internalize steps by removing CoT tokens. 

\section{\label{sec:limitations}Limitations}
\paragraph{Training costs.}
One limitation of the proposed approach is its high training cost due to the finetuning required when removing each set of CoT tokens. As discussed in \Cref{sec:ablation}, removing CoT tokens too fast leads to non-convergence. Therefore, the longer the CoT chain, the longer the training duration. For tasks like $N$-digit multiplication, where the reasoning chain length grows exponentially with $N$, training becomes expensive as $N$ increases. 

\paragraph{Instability.}
Another practical issue we observed is the instability of training with aggressive $\Delta$ values. For example, \Cref{fig:instability} in \Cref{sec:instability} shows a case where the model could not recover from a drop in accuracy. Using lower $\Delta$ values generally leads to more stable training, but at the cost of longer training time. Identifying and addressing unstable dynamics early on, potentially by restarting training as suggested by \citet{hu2024latent}, could be a valuable improvement.

\paragraph{Interpretability.}
Similar to existing work on No CoT and implicit CoT training, models trained using our approach lose interpretable intermediate steps. However, it might be possible to interpret the internal hidden states of these models using probing techniques~\citep{belinkov2018internal,hewitt2019designing}. Additionally, combining implicit and explicit CoT training could allow users to choose between interpretability and latency, providing flexibility based on the requirements of future tasks.

\paragraph{Accuracy.}
Undoubtedly, explicit CoT still achieves higher accuracies compared to our approach to implicit CoT. However, our method enables a trade-off between latency and accuracy. Even on tasks it cannot fully solve without intermediate steps, such as $11\times11$ multiplication, it maintains reasonable accuracy while being several times faster than explicit CoT. Moreover, our results demonstrate the potential of leveraging hidden states for reasoning: even a GPT-2 Small model can be trained to solve $9\times9$ multiplication, despite having only 12 layers, far fewer than the number of reasoning steps in the CoT for $9\times9$ multiplication. When scaled to larger models with hundreds of billions of parameters and up to a hundred layers, such as GPT-3~\citep{brown2020language}, they could potentially solve even more challenging reasoning tasks without explicit CoT steps.

\section{Conclusions and Future Work}
In this work, we introduced Stepwise Internalization, a novel approach for achieving implicit chain-of-thought reasoning in language models. By gradually removing intermediate CoT tokens and finetuning the model, we enable the internalization of reasoning steps incrementally. Our approach demonstrates significant improvements over existing methods, achieving high accuracy on up to $9\times9$ multiplication using GPT-2 Small and outperforming GPT-4 on GSM8K while not outputting any intermediate reasoning steps. Compared to explicit CoT methods, our approach can be up to 11 times faster while maintaining similar accuracies.

For future work, probing the internal processes as the model internalizes each reasoning step could provide insights into the learning mechanisms. Additionally, developing a mixed-mode approach that combines implicit and explicit CoT reasoning could potentially offer the best of both worlds, balancing accuracy, latency, and interpretability based on user preferences. Another promising direction is scaling Stepwise Internalization to larger models and more extensive training/pretraining setups, which could further enhance its effectiveness on a broader range of reasoning tasks.

\begin{ack}
This work was supported by NSF grant DMS-2134012 and ONR grant N00014-24-1-2207. We also thank Harvard University FAS Research Computing for providing computational resources.
\end{ack}

\bibliography{neurips_2024}

\begin{thebibliography}{20}
\providecommand{\natexlab}[1]{#1}
\providecommand{\url}[1]{\texttt{#1}}
\expandafter\ifx\csname urlstyle\endcsname\relax
  \providecommand{\doi}[1]{doi: #1}\else
  \providecommand{\doi}{doi: \begingroup \urlstyle{rm}\Url}\fi

\bibitem[Abdin et~al.(2024)Abdin, Jacobs, Awan, Aneja, Awadallah, Awadalla, Bach, Bahree, Bakhtiari, Behl, Benhaim, Bilenko, Bjorck, Bubeck, Cai, Mendes, Chen, Chaudhary, Chopra, Giorno, de~Rosa, Dixon, Eldan, Iter, Garg, Goswami, Gunasekar, Haider, Hao, Hewett, Huynh, Javaheripi, Jin, Kauffmann, Karampatziakis, Kim, Khademi, Kurilenko, Lee, Lee, Li, Liang, Liu, Lin, Lin, Madan, Mitra, Modi, Nguyen, Norick, Patra, Perez-Becker, Portet, Pryzant, Qin, Radmilac, Rosset, Roy, Ruwase, Saarikivi, Saied, Salim, Santacroce, Shah, Shang, Sharma, Song, Tanaka, Wang, Ward, Wang, Witte, Wyatt, Xu, Xu, Yadav, Yang, Yang, Yu, Zhang, Zhang, Zhang, Zhang, Zhang, Zhang, Zhang, and Zhou]{abdin2024phi3}
Marah Abdin, Sam~Ade Jacobs, Ammar~Ahmad Awan, Jyoti Aneja, Ahmed Awadallah, Hany Awadalla, Nguyen Bach, Amit Bahree, Arash Bakhtiari, Harkirat Behl, Alon Benhaim, Misha Bilenko, Johan Bjorck, Sébastien Bubeck, Martin Cai, Caio César~Teodoro Mendes, Weizhu Chen, Vishrav Chaudhary, Parul Chopra, Allie~Del Giorno, Gustavo de~Rosa, Matthew Dixon, Ronen Eldan, Dan Iter, Amit Garg, Abhishek Goswami, Suriya Gunasekar, Emman Haider, Junheng Hao, Russell~J. Hewett, Jamie Huynh, Mojan Javaheripi, Xin Jin, Piero Kauffmann, Nikos Karampatziakis, Dongwoo Kim, Mahoud Khademi, Lev Kurilenko, James~R. Lee, Yin~Tat Lee, Yuanzhi Li, Chen Liang, Weishung Liu, Eric Lin, Zeqi Lin, Piyush Madan, Arindam Mitra, Hardik Modi, Anh Nguyen, Brandon Norick, Barun Patra, Daniel Perez-Becker, Thomas Portet, Reid Pryzant, Heyang Qin, Marko Radmilac, Corby Rosset, Sambudha Roy, Olatunji Ruwase, Olli Saarikivi, Amin Saied, Adil Salim, Michael Santacroce, Shital Shah, Ning Shang, Hiteshi Sharma, Xia Song, Masahiro Tanaka, Xin Wang, Rachel
  Ward, Guanhua Wang, Philipp Witte, Michael Wyatt, Can Xu, Jiahang Xu, Sonali Yadav, Fan Yang, Ziyi Yang, Donghan Yu, Chengruidong Zhang, Cyril Zhang, Jianwen Zhang, Li~Lyna Zhang, Yi~Zhang, Yue Zhang, Yunan Zhang, and Xiren Zhou.
\newblock Phi-3 technical report: A highly capable language model locally on your phone, 2024.

\bibitem[Belinkov(2018)]{belinkov2018internal}
Yonatan Belinkov.
\newblock \emph{On internal language representations in deep learning: An analysis of machine translation and speech recognition}.
\newblock PhD thesis, Massachusetts Institute of Technology, 2018.

\bibitem[bench authors(2023)]{srivastava2023beyond}
BIG bench authors.
\newblock Beyond the imitation game: Quantifying and extrapolating the capabilities of language models.
\newblock \emph{Transactions on Machine Learning Research}, 2023.
\newblock ISSN 2835-8856.
\newblock URL \url{https://openreview.net/forum?id=uyTL5Bvosj}.

\bibitem[Brown et~al.(2020)Brown, Mann, Ryder, Subbiah, Kaplan, Dhariwal, Neelakantan, Shyam, Sastry, Askell, Agarwal, Herbert-Voss, Krueger, Henighan, Child, Ramesh, Ziegler, Wu, Winter, Hesse, Chen, Sigler, Litwin, Gray, Chess, Clark, Berner, McCandlish, Radford, Sutskever, and Amodei]{brown2020language}
Tom~B. Brown, Benjamin Mann, Nick Ryder, Melanie Subbiah, Jared Kaplan, Prafulla Dhariwal, Arvind Neelakantan, Pranav Shyam, Girish Sastry, Amanda Askell, Sandhini Agarwal, Ariel Herbert-Voss, Gretchen Krueger, Tom Henighan, Rewon Child, Aditya Ramesh, Daniel~M. Ziegler, Jeffrey Wu, Clemens Winter, Christopher Hesse, Mark Chen, Eric Sigler, Mateusz Litwin, Scott Gray, Benjamin Chess, Jack Clark, Christopher Berner, Sam McCandlish, Alec Radford, Ilya Sutskever, and Dario Amodei.
\newblock Language models are few-shot learners, 2020.

\bibitem[Cobbe et~al.(2021)Cobbe, Kosaraju, Bavarian, Chen, Jun, Kaiser, Plappert, Tworek, Hilton, Nakano, Hesse, and Schulman]{cobbe2021training}
Karl Cobbe, Vineet Kosaraju, Mohammad Bavarian, Mark Chen, Heewoo Jun, Lukasz Kaiser, Matthias Plappert, Jerry Tworek, Jacob Hilton, Reiichiro Nakano, Christopher Hesse, and John Schulman.
\newblock Training verifiers to solve math word problems, 2021.

\bibitem[Deng et~al.(2023)Deng, Prasad, Fernandez, Smolensky, Chaudhary, and Shieber]{deng2023implicit}
Yuntian Deng, Kiran Prasad, Roland Fernandez, Paul Smolensky, Vishrav Chaudhary, and Stuart Shieber.
\newblock Implicit chain of thought reasoning via knowledge distillation, 2023.

\bibitem[Dziri et~al.(2024)Dziri, Lu, Sclar, Li, Jiang, Lin, Welleck, West, Bhagavatula, Le~Bras, et~al.]{dziri2024faith}
Nouha Dziri, Ximing Lu, Melanie Sclar, Xiang~Lorraine Li, Liwei Jiang, Bill~Yuchen Lin, Sean Welleck, Peter West, Chandra Bhagavatula, Ronan Le~Bras, et~al.
\newblock Faith and fate: Limits of transformers on compositionality.
\newblock \emph{Advances in Neural Information Processing Systems}, 36, 2024.

\bibitem[Hewitt and Liang(2019)]{hewitt2019designing}
John Hewitt and Percy Liang.
\newblock Designing and interpreting probes with control tasks, 2019.

\bibitem[Hu et~al.(2024)Hu, Chen, Saphra, and Cho]{hu2024latent}
Michael~Y. Hu, Angelica Chen, Naomi Saphra, and Kyunghyun Cho.
\newblock Latent state models of training dynamics, 2024.

\bibitem[Jiang et~al.(2023)Jiang, Sablayrolles, Mensch, Bamford, Chaplot, de~las Casas, Bressand, Lengyel, Lample, Saulnier, Lavaud, Lachaux, Stock, Scao, Lavril, Wang, Lacroix, and Sayed]{jiang2023mistral}
Albert~Q. Jiang, Alexandre Sablayrolles, Arthur Mensch, Chris Bamford, Devendra~Singh Chaplot, Diego de~las Casas, Florian Bressand, Gianna Lengyel, Guillaume Lample, Lucile Saulnier, Lélio~Renard Lavaud, Marie-Anne Lachaux, Pierre Stock, Teven~Le Scao, Thibaut Lavril, Thomas Wang, Timothée Lacroix, and William~El Sayed.
\newblock Mistral 7b, 2023.

\bibitem[Kingma and Ba(2017)]{kingma2017adam}
Diederik~P. Kingma and Jimmy Ba.
\newblock Adam: A method for stochastic optimization, 2017.

\bibitem[Lehnert et~al.(2024)Lehnert, Sukhbaatar, Su, Zheng, Mcvay, Rabbat, and Tian]{lehnert2024a}
Lucas Lehnert, Sainbayar Sukhbaatar, DiJia Su, Qinqing Zheng, Paul Mcvay, Michael Rabbat, and Yuandong Tian.
\newblock Beyond a*: Better planning with transformers via search dynamics bootstrapping, 2024.

\bibitem[Loshchilov and Hutter(2019)]{loshchilov2018decoupled}
Ilya Loshchilov and Frank Hutter.
\newblock Decoupled weight decay regularization.
\newblock In \emph{International Conference on Learning Representations}, 2019.
\newblock URL \url{https://openreview.net/forum?id=Bkg6RiCqY7}.

\bibitem[Nye et~al.(2021)Nye, Andreassen, Gur-Ari, Michalewski, Austin, Bieber, Dohan, Lewkowycz, Bosma, Luan, Sutton, and Odena]{nye2021work}
Maxwell Nye, Anders~Johan Andreassen, Guy Gur-Ari, Henryk Michalewski, Jacob Austin, David Bieber, David Dohan, Aitor Lewkowycz, Maarten Bosma, David Luan, Charles Sutton, and Augustus Odena.
\newblock Show your work: Scratchpads for intermediate computation with language models, 2021.

\bibitem[Pfau et~al.(2024)Pfau, Merrill, and Bowman]{pfau2024lets}
Jacob Pfau, William Merrill, and Samuel~R. Bowman.
\newblock Let's think dot by dot: Hidden computation in transformer language models, 2024.

\bibitem[Radford et~al.(2019)Radford, Wu, Child, Luan, Amodei, Sutskever, et~al.]{radford2019language}
Alec Radford, Jeffrey Wu, Rewon Child, David Luan, Dario Amodei, Ilya Sutskever, et~al.
\newblock Language models are unsupervised multitask learners.
\newblock \emph{OpenAI blog}, 1\penalty0 (8):\penalty0 9, 2019.

\bibitem[Shen et~al.(2023)Shen, Bubeck, Eldan, Lee, Li, and Zhang]{shen2023positional}
Ruoqi Shen, Sébastien Bubeck, Ronen Eldan, Yin~Tat Lee, Yuanzhi Li, and Yi~Zhang.
\newblock Positional description matters for transformers arithmetic, 2023.

\bibitem[Snell et~al.(2022)Snell, Klein, and Zhong]{snell2022learning}
Charlie Snell, Dan Klein, and Ruiqi Zhong.
\newblock Learning by distilling context, 2022.

\bibitem[Wei et~al.(2022)Wei, Wang, Schuurmans, Bosma, brian ichter, Xia, Chi, Le, and Zhou]{wei2022chain}
Jason Wei, Xuezhi Wang, Dale Schuurmans, Maarten Bosma, brian ichter, Fei Xia, Ed~H. Chi, Quoc~V Le, and Denny Zhou.
\newblock Chain of thought prompting elicits reasoning in large language models.
\newblock In Alice~H. Oh, Alekh Agarwal, Danielle Belgrave, and Kyunghyun Cho, editors, \emph{Advances in Neural Information Processing Systems}, 2022.
\newblock URL \url{https://openreview.net/forum?id=_VjQlMeSB_J}.

\bibitem[Yang et~al.(2023)Yang, Ding, Lv, Jiang, He, Guo, Bai, and Tang]{yang2023gpt}
Zhen Yang, Ming Ding, Qingsong Lv, Zhihuan Jiang, Zehai He, Yuyi Guo, Jinfeng Bai, and Jie Tang.
\newblock Gpt can solve mathematical problems without a calculator, 2023.

\end{thebibliography}
\bibliographystyle{plainnat}


\newpage
\appendix

\section{\label{sec:hyperparameters}Hyperparameters}
For all experiments, we use the AdamW optimizer~\citep{loshchilov2018decoupled}, with $\lambda=4$ and an effective batch size of 32 by default. For Phi-3 3.8B and Mistral 7B, we use a batch size of 16 with a gradient accumulation of 2. For the multiplication tasks, we use a learning rate of $5 \times 10^{-5}$ and $\Delta=8$. For GSM8K, we use a learning rate of $5 \times 10^{-5}$ and $\Delta=1$ for GPT-2 Small and GPT-2 Medium, and a learning rate of $1 \times 10^{-5}$ and $\Delta=8$ for Phi-3 3.8B and Mistral 7B, with bfloat16 precision. Additionally, for GSM8K, we only consider sequences with 150 or fewer tokens for training and remove all CoT tokens when 39 or more tokens are scheduled to be removed. All experiments are run on a single H100 with 80GB of GPU memory for up to 200 epochs or 24 hours, whichever is reached first.

\section{\label{sec:instability}Stability Issues for Aggressive Removal}
\begin{figure}[!t]
  \centering
  \includegraphics[width=0.98\textwidth]{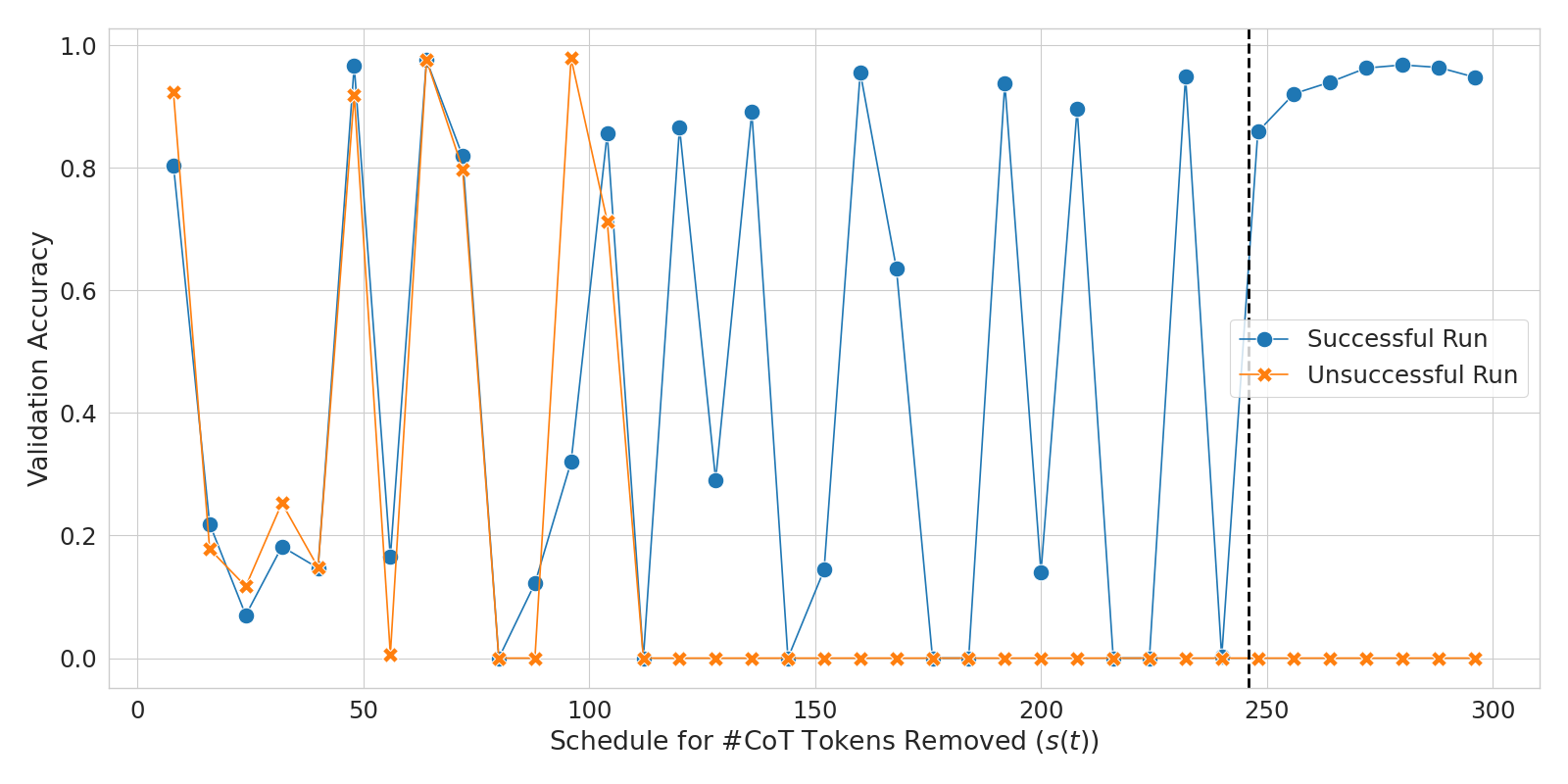}
  \caption{\label{fig:instability}Validation Accuracy during Training for two different random seeds. This figure plots the validation accuracy as a function of the potential number of removed CoT tokens during training for the $9\times9$ multiplication task using GPT-2 Small and $\Delta=8$. The two curves only differ in random seeds. The black dashed vertical line indicates the point beyond which all CoT tokens are removed.}
\end{figure}
We found that using aggressive removal schedules (that is, bigger $\Delta$ values) can sometimes lead to unstable training dynamics. As one example, \Cref{fig:instability} shows two different runs under identical configurations except for the random seed. One run was eventually able to solve the task after all CoT tokens were removed, whereas the other failed to solve the task after all CoT tokens were removed.

\section{Additional Experiments}
\paragraph{Keeping position IDs.} As CoT tokens are removed, the position where the final output starts changes. We tried a variant where position IDs remain unchanged, meaning the position ID of the next token is used directly after removing a CoT token. Although this approach was more stable during training, its performance was similar to the current approach. For simplicity, we did not use this variant in our main experiments.

\paragraph{Alternative CoT formats.}
Different valid reasoning paths can lead to the correct final answer for the same problem. We explored using a binary tree formatted CoT chain for the multiplication problems. This format decomposes an $N$-digit multiplication into a sequence of $N$-digit-by-1-digit multiplication problems, merges the results using sum operators, and continues merging until the final sum is computed. This program has a shorter description length, potentially making it easier for transformers to learn~\citep{dziri2024faith}. However, its performance was similar to the current approach: for $9\times9$ multiplication using GPT-2 Small, it achieved 0.95 accuracy and failed on $11\times11$ multiplication. 




\end{document}